\documentclass[11pt]{article}

\PassOptionsToPackage{hyperfootnotes=false}{hyperref}
\usepackage[final]{acl}

\usepackage{times}
\usepackage{latexsym}

\usepackage[T1]{fontenc}

\usepackage[utf8]{inputenc}

\usepackage{microtype}

\usepackage{inconsolata}

\usepackage{graphicx}

\usepackage{amsmath,amssymb,amsthm}
\usepackage{booktabs}
\usepackage{multirow}
\usepackage{nicefrac}
\usepackage{pifont}

\theoremstyle{plain}

\theoremstyle{definition}

\title{FG$^2$-GDN: Enhancing Long-Context Gated Delta Networks with Doubly Fine-Grained Control}

\author{
  \textbf{Pingwei Sun\textsuperscript{1}},
  \textbf{Yuxuan Hu\textsuperscript{1}},
  \textbf{Jianchao Tan\textsuperscript{1}\thanks{\ Corresponding author.}},
  \textbf{Xue Wang\textsuperscript{1}},
  \textbf{Jiaqi Zhang\textsuperscript{1}},
\\
  \textbf{Yifan Lu\textsuperscript{1}},
  \textbf{Yerui Sun\textsuperscript{1}},
  \textbf{Yuchen Xie\textsuperscript{1}},
  \textbf{Xunliang Cai\textsuperscript{1}}
\\
  Meituan\textsuperscript{1}
\\
  \texttt{\{sunpingwei,tanjianchao02\}@meituan.com}
}

\begin{document}
\maketitle
\begin{abstract}
  Linear attention mechanisms have emerged as promising alternatives to softmax attention, offering linear-time complexity during inference. Recent advances such as Gated DeltaNet (GDN) and Kimi Delta Attention (KDA) have demonstrated that the delta rule (an online gradient descent update) enables superior associative recall compared to simple additive updates.
  While KDA refined the coarse head-wise decay gate into channel-wise decay, the learning rate $\beta_t$ in the delta update remains a scalar, limiting the model's capacity for dimension-specific adaptation.
  We introduce \textbf{FG$^2$-GDN} (Doubly Fine-Grained Gated Delta Network), which replaces the scalar $\beta_t$ with a channel-wise vector in a manner analogous to the transition from SGD to per-coordinate adaptive optimizers such as AdaGrad and Adam.
  We further propose \textbf{FG$^2$-GDN+}, which decouples the scaling for keys and values, allowing value-side write scaling to vary separately from key-side erasure scaling.
  Across 340M and 1.3B models, fine-grained writing improves aggregate language-modeling and long-context results over GDN and KDA. FG$^2$-GDN+ progressively combines channel-wise writing rates with separate key/value control and gives the strongest aggregate RULER and LongBench results. The variants add less than 1\% architectural FLOPs and retain comparable prefill efficiency.
\end{abstract}

\section{Introduction}

The Transformer architecture has revolutionized natural language processing through its powerful attention mechanism, enabling models to capture long-range dependencies effectively. However, the quadratic complexity of softmax attention with respect to sequence length poses significant challenges for both training efficiency and inference scalability, particularly for long-context applications such as agentic and multimodal scenarios.

Linear attention mechanisms \citep{katharopoulos2020transformers} address this limitation by reformulating attention as a linear recurrence with matrix-valued hidden states, reducing inference complexity from $O(L^2)$ to $O(L)$. Early variants of linear attention underperformed softmax attention, but recent innovations have substantially narrowed this gap. Two key developments have driven this progress: (1) \textit{gating mechanisms} that enable adaptive memory decay \citep{hua2022transformer, peng2023rwkv, yang2024gated}, and (2) the \textit{delta rule} that facilitates precise memory updates through online gradient descent \citep{schlag2021linear, yang2024parallelizing}.

The delta rule treats the recurrent state as a learnable associative memory that continuously corrects itself toward the target mapping $k_t \mapsto v_t$. Gated DeltaNet (GDN) \citep{yang2024gated} subsequently unified the delta rule with gating mechanisms, introducing a scalar decay factor $\alpha_t \in (0, 1)$ that enables controlled memory forgetting.

More recently, Kimi Delta Attention (KDA) \citep{team2025kimi} refined GDN's coarse head-wise decay into channel-wise decay through a diagonal gating matrix $\text{Diag}(\alpha_t)$, which enables each feature dimension to maintain independent forgetting rates. KDA demonstrated that such fine-grained control significantly improves long-context modeling and associative recall.

\textbf{Our Contribution.} In this work, we propose the \emph{Doubly Fine-Grained Gated Delta Network} (FG$^2$-GDN), a complementary improvement to the KDA framework. While KDA introduced channel-wise decay gates, the learning rate $\beta_t$ in the delta update remains a scalar shared across all dimensions. This design choice limits the model's ability to perform dimension-specific adaptation.

We propose \textbf{FG$^2$-GDN}, which extends KDA by introducing fine-grained adaptive learning rates. Specifically, FG$^2$-GDN replaces the scalar $\beta_t \in \mathbb{R}$ with a vector $\beta_t \in \mathbb{R}^{d_k}$, enabling each dimension to maintain independent adaptation dynamics. The update rule transforms as follows:
\begin{align}
S_t &= (I - \tilde{k}_t \tilde{k}_t^\top) \text{Diag}(\alpha_t) S_{t-1} + \tilde{k}_t \tilde{v}_t^\top \nonumber\\
o_t &= S^\top_t q_t
\end{align}
where $\tilde{k}_t = \sqrt{\beta_t} \odot k_t$ and $\tilde{v}_t = \sqrt{\beta_t} \odot v_t$ are channel-wise scaled versions of the key and value vectors.

From an online learning perspective, this modification corresponds to employing per-coordinate learning rates in stochastic gradient descent, analogous to the distinction between SGD and per-coordinate adaptive methods such as AdaGrad \citep{duchi2011adaptive} and Adam \citep{kingma2014adam}.

Experiments at the 340M and 1.3B scales show that FG$^2$-GDN improves long-context retrieval on RULER and LongBench while maintaining competitive language modeling quality. Because the update rule preserves KDA's tied diagonal-plus-rank-1 transition, our method can directly reuse KDA's specialized chunkwise parallel implementation \citep{yang2024fla}, incurring less than 5\% prefill throughput overhead relative to KDA.

\section{Preliminary}

\subsection{Linear Attention and the Delta Rule}

Linear attention \citep{katharopoulos2020transformers} reformulates the attention mechanism by replacing the exponential kernel in softmax attention with a dot-product kernel. Given query, key, and value vectors $q_t, k_t, v_t \in \mathbb{R}^d$, the output is computed as:
\begin{equation}
o_t = \sum_{i=1}^{t}\frac{ \phi(q_t)^\top \phi(k_i) v_i}{\sum_{i=1}^{t} \phi(q_t)^\top \phi(k_i)}
\end{equation}
where $\phi: \mathbb{R}^d \to \mathbb{R}^n$ is a feature map. By setting $\phi$ to the identity mapping and omitting the normalization denominator, we obtain the simplified linear attention recurrence:
\begin{equation}
S_t = S_{t-1} + k_t v_t^\top, \quad o_t = S_t^\top q_t
\label{eq:linear_attn}
\end{equation}
where $S_t \in \mathbb{R}^{d \times d}$ is the matrix-valued hidden state that accumulates key-value associations.

\paragraph{Online Learning Interpretation.} From the fast-weight perspective \citep{schmidhuber1992learning, irie2021going}, $S_t$ serves as an associative memory storing transient mappings from keys to values. The update in Equation~\ref{eq:linear_attn} can be viewed as performing gradient descent on the unbounded correlation objective:

\begin{equation}
\mathcal{L}_t(S) = -\langle S^\top k_t, v_t \rangle
\end{equation}

which continually reinforces recent key-value pairs without any forgetting mechanism. This can cause unbounded growth in the norm of the accumulated state and increasing interference over long contexts; the state shape itself remains fixed.

\paragraph{DeltaNet.} DeltaNet \citep{schlag2021linear} addresses this limitation by interpreting the recurrence as online gradient descent on a reconstruction objective:
\begin{equation}
\mathcal{L}_t(S) = \frac{1}{2} \|S^\top k_t - v_t\|^2
\end{equation}
Taking a gradient step with learning rate $\beta_t$ yields the delta update rule:
\begin{align}
S_t &= S_{t-1} - \beta_t \nabla_S \mathcal{L}_t(S_{t-1}) \nonumber\\
    &= (I - \beta_t k_t k_t^\top) S_{t-1} + \beta_t k_t v_t^\top
\label{eq:deltanet}
\end{align}
This rule treats $S$ as a learnable associative memory that continuously corrects itself toward the mapping $k_t \mapsto v_t$. Its tied rank-1 transition admits a hardware-efficient WY-based chunkwise parallelization \citep{yang2024parallelizing}.

\subsection{Delta Rule with Forgetting Gates}

DeltaNet retains all past associations indefinitely, which can lead to interference over long contexts. Gated DeltaNet (GDN) \citep{yang2024gated} addresses this by introducing a scalar forget gate $\alpha_t \in (0, 1)$:
\begin{equation}
S_t = \alpha_t (I - \beta_t k_t k_t^\top) S_{t-1} + \beta_t k_t v_t^\top
\label{eq:gdn}
\end{equation}
where $\alpha_t$ acts as a data-dependent decay on the hidden state, providing explicit control over memory lifespan. In GDN, both $\alpha_t$ and $\beta_t$ are scalars shared across all feature dimensions: every dimension forgets at the same rate and writes with the same strength.

Kimi Delta Attention (KDA) \citep{team2025kimi} refines this design by replacing the scalar $\alpha_t$ with a channel-wise decay vector $\alpha_t \in \mathbb{R}^{d_k}$:
\begin{equation}
S_t = (I - \beta_t k_t k_t^\top) \text{Diag}(\alpha_t) S_{t-1} + \beta_t k_t v_t^\top
\label{eq:kda}
\end{equation}
where $\text{Diag}(\alpha_t) \in \mathbb{R}^{d_k \times d_k}$ is the diagonal matrix with $\alpha_t$ on its diagonal. This allows each feature dimension to maintain an independent forgetting rate, giving the model finer control over which information to retain or discard. However, the learning rate $\beta_t$ in the delta update remains a scalar, meaning all dimensions still share the same writing strength.

\paragraph{Relation to DPLR.} Both GDN and KDA fit the Diagonal-Plus-Low-Rank (DPLR) transition structure \citep{gu2021efficiently}:
\begin{equation}
S_t = (D_t - a_t b_t^\top) S_{t-1} + c_t v_t^\top.
\end{equation}
For KDA, $D_t = \text{Diag}(\alpha_t)$, $a_t = \beta_t k_t$, $b_t = k_t \odot \alpha_t$, and $c_t = \beta_t k_t$; GDN is the scalar-diagonal special case. General DPLR recurrences admit chunkwise parallelization, while the additional factor tying inherited from the delta rule enables the specialized WY-based algorithm used by GDN and KDA \citep{yang2024parallelizing, team2025kimi}.

\subsection{Per-Coordinate Adaptive Learning Rates}
\label{sec:prelim-adaptive}

The idea of assigning different learning rates to different parameters has a long history in optimization. AdaGrad \citep{duchi2011adaptive} accumulates squared gradients to scale each coordinate inversely by its historical magnitude, enabling faster progress along infrequent-gradient dimensions. Adam \citep{kingma2014adam} further combines per-coordinate scaling with momentum, and has become the default optimizer for training deep networks. A common insight underlying these methods is that a single global step size is often suboptimal when different coordinates of the parameter space exhibit different curvatures or update frequencies. Per-coordinate adaptation can yield faster and more stable convergence in such settings. This observation motivates our extension of the delta rule's scalar learning rate to a channel-wise vector, as described in the next section.

\section{Gated Delta Net with Fine-grained Beta}

\subsection{Motivation}

As discussed in \S\ref{sec:prelim-adaptive}, per-coordinate adaptive learning rates have proven broadly effective in optimization. A similar asymmetry exists in the gated delta rule: KDA introduced channel-wise forgetting ($\alpha_t \in \mathbb{R}^{d_k}$) but retained a scalar writing rate ($\beta_t \in \mathbb{R}$). This means all dimensions share the same learning rate during each delta update, regardless of their individual relevance to the current context. We propose to close this gap by extending $\beta_t$ to a channel-wise vector as well.

\subsection{\texorpdfstring{FG$^2$-GDN}{FG2-GDN} Update Rule}

A straightforward way to introduce per-dimension learning rates would be to replace $\beta_t k_t k_t^\top$ in Equation~\ref{eq:kda} with $\text{Diag}(\beta_t)\, k_t k_t^\top$, where $\beta_t \in \mathbb{R}^{d_k}$. Although $\text{Diag}(\beta_t)\, k_t k_t^\top = (\beta_t \odot k_t) k_t^\top$ remains rank-1, its left and right factors are no longer tied. It therefore falls outside the symmetric Householder-style form used by KDA's specialized WY-based chunkwise algorithm. RWKV-7 \citep{peng2025rwkv7} employs a related untied rank-1 transition and accordingly uses a more general chunkwise parallelization strategy.

To preserve compatibility with the efficient chunkwise parallel framework, we instead absorb $\beta_t$ symmetrically into both $k_t$ and $v_t$. Concretely, we define the scaled key and value as
\begin{align}
\tilde{k}_t = \sqrt{\beta_t} \odot k_t, \quad \tilde{v}_t = \sqrt{\beta_t} \odot v_t
\end{align}
and write the FG$^2$-GDN update as:
\begin{equation}
S_t = (I - \tilde{k}_t \tilde{k}_t^\top)\, \text{Diag}(\alpha_t)\, S_{t-1} + \tilde{k}_t \tilde{v}_t^\top
\label{eq:fgbkda}
\end{equation}
where $\odot$ denotes element-wise multiplication and $\sqrt{\cdot}$ is applied element-wise. The delta subtraction term is now $\tilde{k}_t \tilde{k}_t^\top = \text{Diag}(\sqrt{\beta_t})\, k_t k_t^\top\, \text{Diag}(\sqrt{\beta_t})$, a \emph{symmetric} rank-1 matrix that preserves KDA's tied Householder-style parameterization. This ensures that the chunkwise parallel algorithm of KDA can be reused (see \S\ref{sec:chunkwise}), while each dimension obtains a distinct effective learning rate.

\subsection{Online Learning Interpretation}

Following the online learning view of DeltaNet \citep{schlag2021linear}, the delta update can be seen as a gradient step on the reconstruction loss $\mathcal{L}_t(S) = \frac{1}{2}\|S^\top k_t - v_t\|^2$. In KDA, this step uses a single scalar $\beta_t$ as the learning rate. FG$^2$-GDN generalizes this to per-coordinate learning rates:
\begin{equation}
\begin{split}
& S \leftarrow S - \eta_t \odot (k_t k^\top_t S - k_t v_t^\top) \\
& \eta_t = \sqrt{\beta_t} \sqrt{\beta_t}^\top, \quad \eta_t \in \mathbb{R}^{d\times d}
\end{split}
\end{equation}
Combined with the decay term $\text{Diag}(\alpha_t)$, this yields Equation~\ref{eq:fgbkda}. The element-wise learning-rate matrix is $\eta_t = \sqrt{\beta_t}\sqrt{\beta_t}^\top$, a rank-1 parameterization. Its Hadamard product with $k_t k_t^\top$ yields the symmetric correction $\text{Diag}(\sqrt{\beta_t})\,k_t k_t^\top\,\text{Diag}(\sqrt{\beta_t})$, rather than the asymmetric correction $\text{Diag}(\beta_t)\,k_t k_t^\top$. This structured coupling preserves the tied factorization used by KDA's specialized parallel algorithm; it is not a free full-matrix learning-rate parameterization.

\subsection{\texorpdfstring{FG$^2$-GDN+}{FG2-GDN+} Variant}

We further consider a variant, \textbf{FG$^2$-GDN+}, that decouples the channel-wise scaling for keys and values by introducing separate vectors $\beta^k_t \in \mathbb{R}^{d_k}$ and $\beta^v_t \in \mathbb{R}^{d_v}$:
\begin{align}
S_t &= (I - \hat{k}_t \hat{k}_t^\top)\,\text{Diag}(\alpha_t)\,S_{t-1} + \hat{k}_t \tilde{v}_t^\top \label{eq:fgbkda_plus} \\
\hat{k}_t &= \sqrt{\beta^k_t} \odot k_t, \quad \tilde{v}_t = \sqrt{\beta^v_t} \odot v_t \nonumber
\end{align}
In FG$^2$-GDN, the same $\beta_t$ scales both $k_t$ and $v_t$, tying the ``erasure strength'' and ``write strength'' together. FG$^2$-GDN+ relaxes this constraint: $\beta^k_t$ controls erasure and the key-side factor of the write term, while $\beta^v_t$ separately controls its value-side scaling. This adds one additional projection head while preserving the same update structure.

\subsection{Gate Parameterization}
\label{sec:gate-parameterization}

Let $x_t\in\mathbb{R}^{D}$ be the layer input, $H$ the number of linear-attention heads, and $d_k=d_v=128$ the head dimensions. We inherit KDA's decay pathway unchanged: a two-layer low-rank projection $f_{\mathrm{proj}}:\mathbb{R}^{D}\!\to\!\mathbb{R}^{128}\!\to\!\mathbb{R}^{H d_k}$ supplies channel-wise inputs to the standard KDA gated-decay transform, yielding $\alpha_t\in(0,1)^{H\times d_k}$. For FG$^2$-GDN, a separate two-layer projection $b_{\mathrm{proj}}:\mathbb{R}^{D}\!\to\!\mathbb{R}^{128}\!\to\!\mathbb{R}^{H d_k}$ followed by a sigmoid produces $\beta_t\in(0,1)^{H\times d_k}$. FG$^2$-GDN+ changes only the last layer to output $H(d_k+d_v)$ logits, which are split and sigmoid-activated into $\beta^k_t$ and $\beta^v_t$. Queries and keys are L2-normalized before the square-root scaling in Equations~\ref{eq:fgbkda}--\ref{eq:fgbkda_plus}.

At the 1.3B scale, FG$^2$-GDN and FG$^2$-GDN+ add only 8.8M (0.62\%) and 13.6M (0.96\%) parameters over KDA, respectively, making clear that their gains do not come from a substantial increase in model size.

\subsection{Chunkwise Parallel Training}
\label{sec:chunkwise}

A key practical advantage of FG$^2$-GDN is that its update rule retains the DPLR form. Specifically, Equation~\ref{eq:fgbkda} can be rewritten as
\begin{equation}
S_t = (\text{Diag}(\alpha_t) + a_t b_t^\top)\,S_{t-1} + \tilde{k}_t \tilde{v}_t^\top
\end{equation}
where $a_t = -\tilde{k}_t$ and $b_t = \tilde{k}_t \odot \alpha_t$. Since $a_t$ and $b_t$ are both derived from $\tilde{k}_t = \sqrt{\beta_t}\odot k_t$, the transition remains a diagonal-plus-rank-1 matrix, the same structural class as KDA.

This structural equivalence means that FG$^2$-GDN can directly reuse KDA's chunkwise-parallel algorithm, which splits the sequence into chunks of size $C$ and applies the WY representation \citep{bischof1987wy} together with the UT transform \citep{joffrain2006accumulating} to efficiently accumulate the tied diagonal-plus-rank-1 transitions within each chunk. The only difference is that the scalar $\beta_t$ in FLA's implementation \citep{yang2024fla} is replaced by the vector $\beta_t \in \mathbb{R}^{d_k}$ when constructing the scaled keys $\tilde{k}_t$. All subsequent matrix operations (the triangular solve for the UT factor, the inter-chunk state propagation, and the intra-chunk causal dot-product) remain identical in form and complexity.

As a result, for sequence length $L$, chunk size $C$, and head dimension $d$, FG$^2$-GDN inherits KDA's $O(Ld^2 + LCd + LC^2)$ training-time complexity; KDA's per-head kernel accounting is $6Ld^2 + 3LCd + LC^2$ \citep{team2025kimi}. The added element-wise square-root and scaling operations are $O(Ld)$; with $d_k=d_v=128$, they contribute approximately $d_k/(d_kd_v)=1/128\approx0.8\%$ of the dominant state-update arithmetic. The extra gate projection is included in the architectural FLOP count, while the measured end-to-end overhead also reflects kernel launches and memory traffic not captured by this element-wise estimate (Section~\ref{sec:efficiency}).

\subsection{Hybrid Architecture}

\begin{figure*}[t]
\centering
\includegraphics[width=1.6\columnwidth]{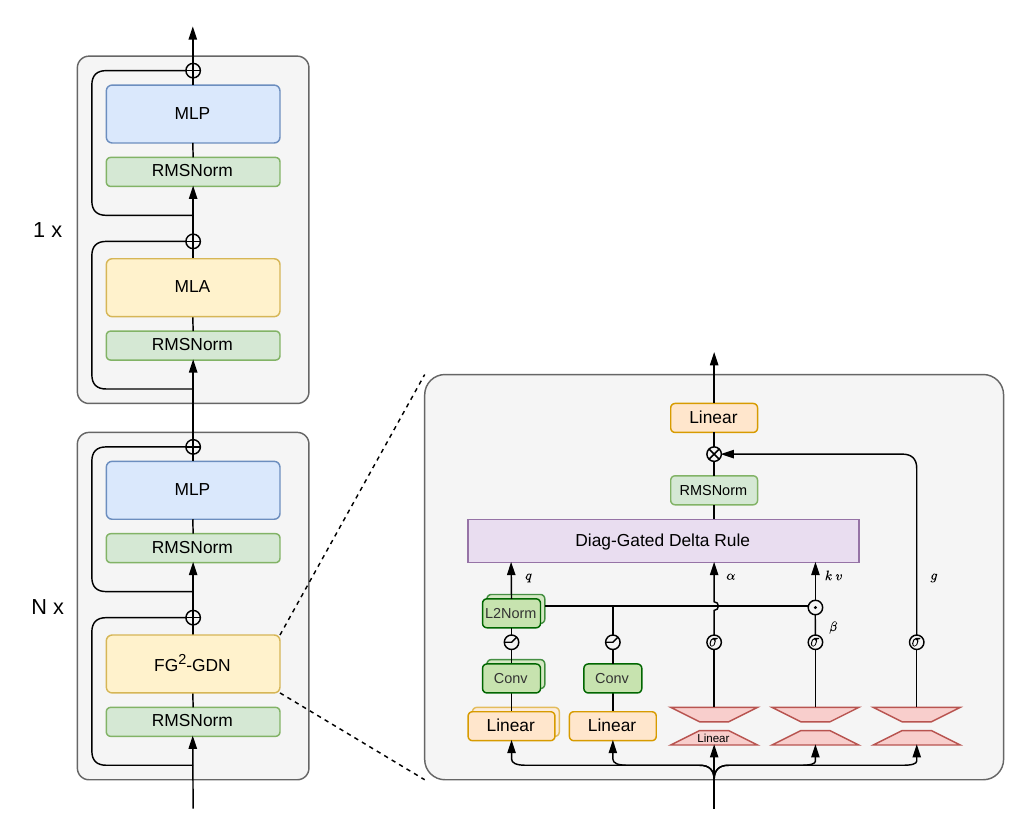}
\caption{Overview of the Linear-MLA hybrid architecture. The backbone follows standard pre-norm residual blocks with SwiGLU MLP. Every $N_{\text{ratio}}$ layers, the linear attention module is replaced by MLA. Within the FG$^2$-GDN module, low-rank projections follow KDA, and the channel-wise $\beta$ pre-scales $k$ and $v$ before the delta update.}
\label{fig:arch}
\end{figure*}

Purely linear attention models are limited by their finite-state capacity \citep{gu2024mamba}. Following recent hybrid designs \citep{yang2024gated, team2025kimi}, we interleave FG$^2$-GDN layers with Multi-head Latent Attention (MLA) \citep{deepseekai2024deepseekv2strongeconomicalefficient} layers at a 3:1 ratio (Figure~\ref{fig:arch}). The MLA layers handle tasks that benefit from full pairwise interaction, while the linear attention layers provide efficient long-range recurrence.

\section{Experiments}

\paragraph{Setup}
Our experiments evaluate FG$^2$-GDN alongside FG$^2$-GDN{+}, which uses separate channel-wise $\beta^k_t$ and $\beta^v_t$ for keys and values, respectively. We compare against two strong baselines: Gated DeltaNet (GDN) \citep{yang2024gated} with scalar $\alpha_t$ and scalar $\beta_t$, and Kimi Delta Attention (KDA) \citep{team2025kimi} with channel-wise $\alpha_t \in \mathbb{R}^{d_k}$ but scalar $\beta_t$. All models use the same 3:1 linear-attention/MLA hybrid backbone, training data, schedule, and hyperparameters; only the linear-attention cell varies. This shared configuration is a controlled comparison rather than per-method tuning. We sample tokens from SlimPajama and train at two scales: \textbf{340M} (15B tokens) and \textbf{1.3B} (100B tokens) using AdamW with an 8K sequence length. Owing to the cost of pretraining at these scales, each configuration is trained once. Full details are in Appendix~\ref{app:exp-details}.

\subsection{Language Modeling}

Table~\ref{tab:lm-benchmarks} summarizes language modeling perplexity and zero-shot accuracy on seven commonsense reasoning benchmarks: LAMBADA \citep{paperno2016lambada}, ARC-Easy/Challenge \citep{clark2018think}, BoolQ \citep{clark2019boolq}, HellaSwag \citep{zellers2019hellaswag}, PIQA \citep{bisk2020piqa}, and Winogrande \citep{sakaguchi2021winogrande}.

\begin{table*}[t]
\centering
\resizebox{\textwidth}{!}{%
\begin{tabular}{cc|c|cccccccc}
\toprule
\multicolumn{2}{c|}{\multirow{2}{*}{\textbf{Model}}} & \textbf{LAMB.} & \textbf{LAMB.} & \textbf{ARC-e} & \textbf{ARC-c} & \textbf{BoolQ} & \textbf{HellaSwag} & \textbf{PIQA} & \textbf{WinoGrande} & \multirow{2}{*}{\textbf{Avg}} \\ \cmidrule{3-10}
\multicolumn{2}{c|}{}                       & ppl   & acc   & acc   & acc\_norm & acc   & acc\_norm & acc   & acc        &                      \\ \midrule
\multirow{4}{*}{340M}       & GDN           & 39.86 & 32.47 & 43.10 & \textbf{24.57}     & 54.77 & 35.75     & 63.82 & 50.12      & 43.51                \\
                            & KDA           & 41.30 & 32.12 & 44.40 & 22.95     & 50.86 & 36.47     & 64.25 & 50.99      & 43.14                \\
                            & FG$^2$-GDN    & 38.27 & \textbf{33.59} & \textbf{45.79} & 22.78     & 58.90 & 36.37     & 64.85 & \textbf{51.85}      & \textbf{44.88}                \\
                            & FG$^2$-GDN+   & \textbf{37.93} & 32.97 & 44.19 & 23.98     & \textbf{59.30} & \textbf{36.87}     & \textbf{65.29} & 51.30      & 44.84                \\ \midrule
\multirow{4}{*}{1.3B}       & GDN           & 14.01 & 48.92 & \textbf{58.63} & 27.99     & 55.69 & 52.51     & 70.73 & 55.09      & 52.79                \\
                            & KDA           & 13.80 & \textbf{48.94} & 56.52 & 26.54     & \textbf{60.43} & \textbf{53.98}     & 71.22 & 54.70      & 53.19                \\
                            & FG$^2$-GDN    & \textbf{13.09} & 48.85 & 57.95 & \textbf{28.92}     & 59.54 & 53.67     & \textbf{71.65} & 57.06      & \textbf{53.95}                \\
                            & FG$^2$-GDN+   & 13.91 & 48.36 & 56.86 & 27.56     & 58.04 & 53.62     & 71.60 & \textbf{57.77}      & 53.40                \\ \bottomrule
\end{tabular}%
}
\caption{Language modeling perplexity (LAMBADA) and zero-shot accuracy (\%) on commonsense reasoning benchmarks at 340M and 1.3B scales.}
\label{tab:lm-benchmarks}
\end{table*}

\textbf{Fine-grained $\beta_t$ improves aggregate language modeling.} Both FG$^2$-GDN and FG$^2$-GDN+ improve average accuracy over KDA at 340M and 1.3B. KDA makes the \emph{forgetting} gate $\alpha_t$ channel-wise, FG$^2$-GDN additionally makes the \emph{writing} rate $\beta_t$ channel-wise, and FG$^2$-GDN+ further decouples its key/value effects. The corrected 1.3B KDA average is 53.19, compared with 53.95 and 53.40 for the two successive variants.

\subsection{Long-Context Evaluation}

We evaluate long-context capability on two benchmarks with the 1.3B models: RULER \citep{hsieh2024ruler}, which tests 13 retrieval categories at sequence lengths 4k--16k (Table~\ref{tab:ruler}), and LongBench \citep{bai2024longbench} (Table~\ref{tab:longbench}), which covers long context understanding tasks like document QA, summarization, few-shot learning, and code completion.

\begin{table*}[t]
\centering
\resizebox{\textwidth}{!}{%
\begin{tabular}{c|c|ccc|ccc|cccccccc}
\toprule
\multirow{2}{*}{\textbf{Model}} &
  \multirow{2}{*}{\textbf{Seq Len}} &
  \multicolumn{3}{c|}{\textbf{NIAH Single}} &
  \multicolumn{3}{c|}{\textbf{NIAH MultiKey}} &
  \multirow{2}{*}{MQ} &
  \multirow{2}{*}{MV} &
  \multirow{2}{*}{CWE} &
  \multirow{2}{*}{FWE} &
  \multirow{2}{*}{QA-H} &
  \multirow{2}{*}{QA-S} &
  \multirow{2}{*}{VT} &
  \multirow{2}{*}{\textbf{Avg}} \\
 &
   &
  S1 &
  S2 &
  S3 &
  MK1 &
  MK2 &
  MK3 &
   &
   &
   &
   &
   &
   &
   &
   \\ \midrule
\multirow{3}{*}{GDN} &
  4k &
  100.0 &
  100.0 &
  86.6 &
  \textbf{94.4} &
  56.4 &
  10.6 &
  78.1 &
  90.2 &
  67.1 &
  36.1 &
  28.2 &
  33.9 &
  19.5 &
  61.6 \\
 &
  8k &
  100.0 &
  100.0 &
  75.0 &
  \textbf{91.4} &
  24.6 &
  3.2 &
  65.8 &
  72.1 &
  36.7 &
  31.1 &
  25.0 &
  23.3 &
  14.2 &
  51.0 \\
 &
  16k &
  99.8 &
  99.0 &
  58.8 &
  71.6 &
  12.6 &
  0.4 &
  41.7 &
  42.4 &
  19.5 &
  58.5 &
  20.0 &
  19.5 &
  \textbf{7.9} &
  42.4 \\ \midrule
\multirow{3}{*}{KDA} &
  4k &
  100.0 &
  100.0 &
  \textbf{93.8} &
  87.6 &
  71.4 &
  \textbf{31.4} &
  77.4 &
  65.5 &
  69.1 &
  43.5 &
  29.8 &
  25.3 &
  4.1 &
  61.5 \\
 &
  8k &
  100.0 &
  100.0 &
  74.4 &
  87.8 &
  51.0 &
  18.2 &
  78.8 &
  67.2 &
  45.7 &
  43.5 &
  25.2 &
  22.1 &
  0.8 &
  55.0 \\
 &
  16k &
  100.0 &
  100.0 &
  74.4 &
  \textbf{74.6} &
  19.6 &
  \textbf{7.8} &
  54.6 &
  35.8 &
  \textbf{26.7} &
  48.7 &
  \textbf{22.0} &
  18.6 &
  1.0 &
  44.9 \\ \midrule
\multirow{3}{*}{FG$^2$-GDN} &
  4k &
  100.0 &
  100.0 &
  89.6 &
  86.0 &
  65.2 &
  14.2 &
  87.3 &
  69.9 &
  \textbf{77.3} &
  \textbf{44.3} &
  28.4 &
  33.6 &
  \textbf{47.1} &
  64.8 \\
 &
  8k &
  100.0 &
  100.0 &
  \textbf{88.2} &
  86.2 &
  38.8 &
  5.0 &
  \textbf{89.1} &
  55.1 &
  \textbf{47.4} &
  \textbf{59.9} &
  27.2 &
  28.2 &
  \textbf{19.6} &
  57.3 \\
 &
  16k &
  100.0 &
  100.0 &
  \textbf{82.6} &
  72.4 &
  26.2 &
  2.4 &
  55.7 &
  27.6 &
  24.2 &
  45.7 &
  19.4 &
  21.1 &
  3.1 &
  44.6 \\ \midrule
\multirow{3}{*}{FG$^2$-GDN+} &
  4k &
  100.0 &
  100.0 &
  88.0 &
  87.8 &
  \textbf{84.6} &
  30.8 &
  \textbf{87.4} &
  \textbf{91.1} &
  70.9 &
  39.3 &
  \textbf{32.8} &
  \textbf{39.3} &
  21.2 &
  \textbf{67.2} \\
 &
  8k &
  100.0 &
  100.0 &
  87.2 &
  79.2 &
  \textbf{75.6} &
  \textbf{19.6} &
  82.1 &
  \textbf{80.6} &
  38.5 &
  55.5 &
  \textbf{27.8} &
  \textbf{33.1} &
  19.4 &
  \textbf{61.4} \\
 &
  16k &
  100.0 &
  98.6 &
  81.2 &
  59.6 &
  \textbf{30.2} &
  4.8 &
  \textbf{68.5} &
  \textbf{54.2} &
  23.2 &
  \textbf{62.8} &
  19.8 &
  \textbf{28.5} &
  3.8 &
  \textbf{48.9} \\
\bottomrule
\end{tabular}%
}
\caption{RULER benchmark results (\%) for 1.3B models at sequence lengths 4k, 8k, and 16k. Categories include single-key NIAH (S1--S3), multi-key NIAH (MK1--MK3), multi-query (MQ), multi-value (MV), common/frequent word extraction (CWE/FWE), question answering (QA-H/QA-S), and variable tracking (VT).}
\label{tab:ruler}
\end{table*}

\begin{table*}[t]
\centering
\resizebox{\linewidth}{!}{%
\begin{tabular}{c|ccc|ccc|ccc|ccc|cc|c}
\toprule
\multirow{2}{*}{\textbf{Model}}
& \multicolumn{3}{c|}{\textbf{Single-Doc QA}}
& \multicolumn{3}{c|}{\textbf{Multi-Doc QA}}
& \multicolumn{3}{c|}{\textbf{Summarization}}
& \multicolumn{3}{c|}{\textbf{Few-shot}}
& \multicolumn{2}{c|}{\textbf{Code}}
& \multirow{2}{*}{\textbf{Avg}} \\

& NQA & QQA & MFQ & HQA & 2WM & Mus & GvR & QMS & MNs & TRC & TQA & SSM & LCC & RBP & \\ \midrule
GDN
& 2.6 & 7.5 & 13.7
& 5.1 & 7.8 & 2.5
& 24.0 & 14.8 & 18.5
& 28.0 & 15.8 & 13.3
& 30.3 & 27.4
& 15.1 \\
KDA
& \textbf{3.7} & 7.8 & \textbf{15.5}
& 4.7 & 7.1 & 3.1
& \textbf{26.9} & 14.2 & 20.3
& 25.0 & 29.1 & 7.8
& \textbf{33.8} & \textbf{30.9}
& 16.4 \\
FG$^2$-GDN
& 3.3 & \textbf{9.2} & 14.1
& 4.9 & 6.8 & 3.4
& 24.5 & 14.6 & 17.8
& 32.5 & 24.6 & 16.0
& 27.3 & 24.9
& 16.0 \\
FG$^2$-GDN+
& 3.1 & 9.1 & 12.5
& \textbf{5.2} & \textbf{7.9} & \textbf{3.8}
& 24.8 & \textbf{15.0} & \textbf{22.0}
& \textbf{45.5} & \textbf{34.5} & \textbf{16.4}
& 30.2 & 26.2
& \textbf{18.3} \\
\bottomrule
\end{tabular}%
}
\caption{LongBench evaluation results for 1.3B models across single-doc QA, multi-doc QA, summarization, few-shot learning, and code completion tasks.}
\label{tab:longbench}
\end{table*}

\textbf{Progressively finer writing control improves long-context memory.} FG$^2$-GDN first introduces a per-coordinate writing rate, which improves several single-key retrieval tasks with varied query patterns (MQ, CWE). FG$^2$-GDN+ then adds independent key/value scaling and achieves the strongest aggregate RULER and LongBench results, including the best MK2 scores across all tested lengths. Individual subtask scores are not monotonic---for example, FG$^2$-GDN is below KDA on overall LongBench (16.0 vs. 16.4), and FG$^2$-GDN+ is below FG$^2$-GDN on 4K variable tracking (21.2 vs. 47.1)---but we do not interpret these separately trained variants as a per-scenario selection menu. They form successive steps along the same writing-granularity axis, with FG$^2$-GDN+ as the complete variant.

\textbf{Graceful extrapolation beyond training length.} Though all models are trained with an 8K sequence length, we extend the test settings to 16K to further explore the extrapolation ability of NoPE-linear hybrid models. As shown in the bottom line of each category, FG$^2$-GDN{+} retains the highest accuracy at 16K, suggesting that fine-grained memory control provides a more structured state representation that better tolerates distribution shift in position.

\subsection{Learned Writing-Rate Analysis}
\label{sec:beta-analysis}

\begin{figure}[t]
\centering
\includegraphics[width=\linewidth]{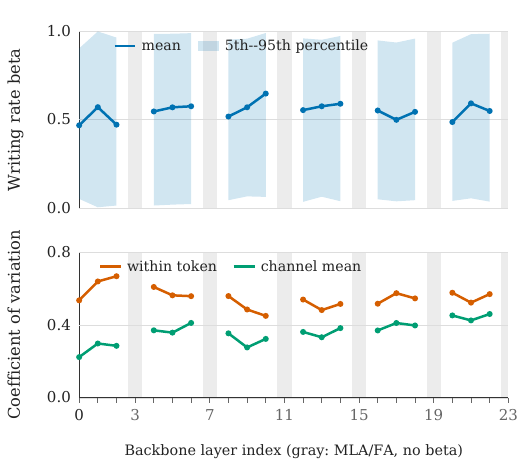}
\caption{Per-layer statistics of the learned $\beta$ in the 1.3B FG$^2$-GDN checkpoint on 1{,}408 held-out tokens. The horizontal axis gives the original 24-layer backbone index; gray bands at layers 3, 7, 11, 15, 19, and 23 mark MLA/full-attention layers, which have no $\beta$ and are skipped. Top: mean and 5th--95th percentile. Bottom: coefficient of variation (CV) within each token and across time-averaged channel means. All 18 linear-attention layers retain substantial channel variation.}
\label{fig:beta-stats}
\end{figure}

Figure~\ref{fig:beta-stats} measures the sigmoid output of $b_{\mathrm{proj}}$ at the 18 linear-attention positions within the 24-layer hybrid backbone, excluding the six interleaved MLA layers. The mean within-token channel CV is 0.55 (layer range 0.45--0.67), and the time-averaged channel means have CV 0.36. Averaged across layers, the 5th and 95th percentiles are 0.04 and 0.96. Thus, for the same token, the model assigns near-zero rates to some coordinates and near-one rates to others---behavior unavailable to a scalar $\beta_t$.

\subsection{Ablation Study}

\begin{table}[t]
\centering
\resizebox{\linewidth}{!}{%
\begin{tabular}{lccc|ccc}
\toprule
\textbf{Model} & \textbf{Vec. $\alpha$} & \textbf{Vec. $\beta$} & \textbf{Sep. $\beta^{k/v}$} & \textbf{LM Avg $\uparrow$} & \textbf{PPL $\downarrow$} & \textbf{LB Avg $\uparrow$} \\
\midrule
GDN & \ding{55} & \ding{55} & \ding{55} & 52.79 & 14.01 & 15.1 \\
FGBGDN & \ding{55} & \ding{51} & \ding{55} & 53.25 & 13.96 & 16.8 \\
KDA & \ding{51} & \ding{55} & \ding{55} & 53.19 & 13.80 & 16.4 \\
\midrule
FG$^2$-GDN & \ding{51} & \ding{51} & \ding{55} & \textbf{53.95} & \textbf{13.09} & 16.0 \\
+ sep. scalar $\beta^{k/v}$ & \ding{51} & \ding{55} & \ding{51} & 53.9 & 13.56 & 16.7 \\
FG$^2$-GDN+ & \ding{51} & \ding{51} & \ding{51} & 53.40 & 13.91 & \textbf{18.3} \\
\bottomrule
\end{tabular}%
}
\caption{Ablation study at 1.3B. FGBGDN is the added isolating cell with scalar decay $\alpha$ and channel-wise writing rate $\beta$. LM Avg is average zero-shot accuracy (\%); LB Avg is the LongBench score (\%).}
\label{tab:ablation}
\end{table}

To disentangle channel-wise decay, channel-wise writing, and separate key/value scaling, we train additional 1.3B variants that enable these axes independently. Table~\ref{tab:ablation} reports language-modeling and LongBench results.

\textbf{Channel-wise $\beta$ helps independently and complements $\alpha$.} Adding vector $\beta$ to the scalar-$\alpha$ GDN base (FGBGDN) improves the 1.3B LM average from 52.79 to 53.25 and LongBench from 15.1 to 16.8. Adding it to the channel-wise-$\alpha$ KDA base further improves LM from 53.19 to 53.95 and lowers LAMBADA perplexity from 13.80 to 13.09. This isolation shows that per-coordinate writing helps independently of, and complements, per-coordinate decay.

\textbf{Separate $\beta^k / \beta^v$ improves long-context tasks.} The variant with separate scalar $\beta^k$ and $\beta^v$ (without vectorization) improves over KDA on both aggregate LM and LongBench metrics, indicating that decoupling erasure and write strength is useful even without channel-wise granularity.

\textbf{The writing-rate refinements are progressive.} FG$^2$-GDN+ combines vectorization and key/value separation, achieving the highest LongBench average (18.3\%). Together with the isolating rows, this supports a progression from scalar writing, to channel-wise writing, and finally to channel-wise key/value control, rather than mutually exclusive variants selected for different tasks.

\subsection{Efficiency}
\label{sec:efficiency}

\begin{figure*}[t]
\centering
\includegraphics[width=0.9\textwidth]{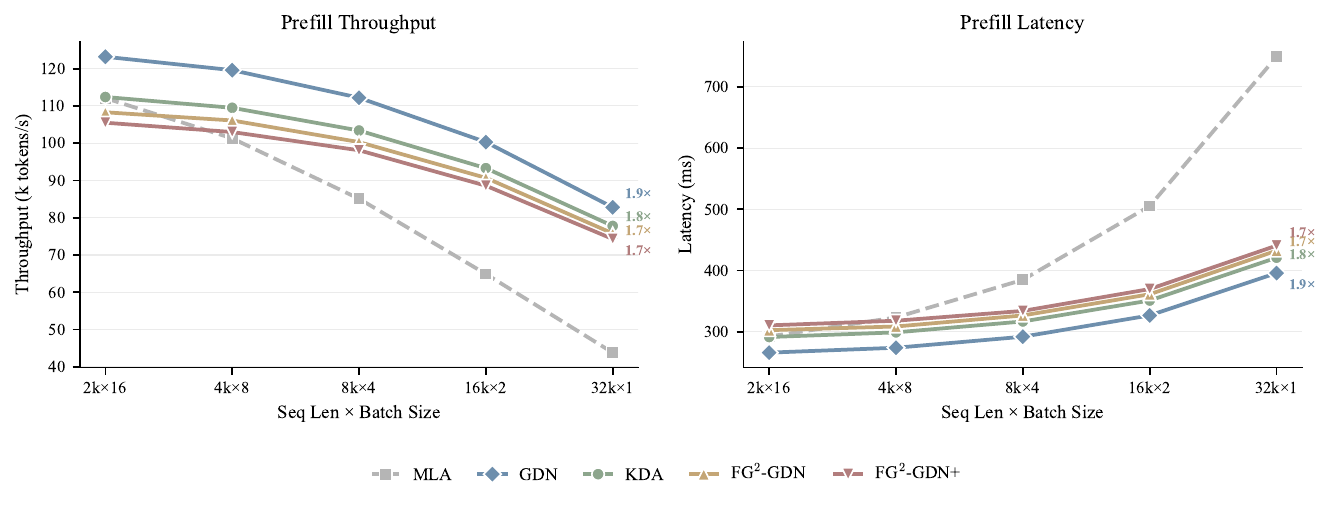}
\caption{Prefill throughput (left) and latency (right) on NVIDIA H800-80G (BF16) with a fixed total token budget of 32{,}768. Numbers at the rightmost point indicate the speedup over MLA at the 32k$\times$1 configuration. MLA (dashed) degrades sharply with increasing sequence length, while all linear attention variants scale similarly.}
\label{fig:prefill-speed}
\end{figure*}

\textbf{Expressiveness at near-zero cost.} Figure~\ref{fig:prefill-speed} compares the prefill throughput and latency across configurations from 2k$\times$16 to 32k$\times$1. At the longest sequence (32k), all linear attention variants achieve 1.7--1.9$\times$ speedup over MLA. The channel-wise $\beta_t$ parameterization introduces less than 5\% overhead relative to KDA across all settings, and the gap narrows as sequences grow, which is precisely the regime where the expressiveness benefits matter most. FG$^2$-GDN{+} incurs a slightly larger overhead from the additional $\beta^v_t$ projection but remains competitive with all baselines.

\textbf{Linear scaling preserved.} The throughput ratio between 32k$\times$1 and 2k$\times$16 is approximately $0.70\times$ for all linear attention variants, compared to $0.39\times$ for MLA, confirming that FG$^2$-GDN inherits the favorable linear-time scaling of the delta-rule family.

\section{Related Work and Discussion}

\paragraph{Linear Attention as Recurrent Memory.}
Linear attention \citep{katharopoulos2020transformers, choromanski2020rethinking} reformulates softmax attention into a linear recurrence with a matrix-valued hidden state $S_t \in \mathbb{R}^{d_k \times d_v}$, which can be viewed as a fast-weight associative memory \citep{schmidhuber1992learning, irie2021going}. A unified framework for these models writes the state transition as $S_t = A_t \, S_{t-1} + k_t v_t^\top$, where $A_t$ is a transition matrix and $o_t = S_t^\top q_t$ reads out the memory. The primary axis of variation across recent models lies in the parameterization of $A_t$ and the form of the write term. Table~\ref{tab:update-rules} summarizes representative methods under this lens.

\begin{table*}[t]
\centering
\resizebox{\textwidth}{!}{%
\begin{tabular}{l|c|c|l}
\toprule
\textbf{Model} & \textbf{Transition $A_t$} & \textbf{Delta rule?} & \textbf{State update} \\
\midrule
Linear Attention \citep{katharopoulos2020transformers} & $I$ & \ding{55} & $S_t = S_{t-1} + k_t v_t^\top$ \\
RetNet \citep{sun2023retentive} & scalar & \ding{55} & $S_t = \gamma \, S_{t-1} + k_t v_t^\top$ \\
Mamba-2 \citep{dao2024transformers} & scalar & \ding{55} & $S_t = \alpha_t \, S_{t-1} + k_t v_t^\top$ \\
GLA \citep{yang2024gla} & diagonal & \ding{55} & $S_t = \text{Diag}(\alpha_t) \, S_{t-1} + k_t v_t^\top$ \\
RWKV-6 \citep{peng2023rwkv} & diagonal & \ding{55} & $S_t = \text{Diag}(\alpha_t) \, S_{t-1} + k_t v_t^\top$ \\
HGRN-2 \citep{qin2024hgrn2} & diagonal & \ding{55} & $S_t = \text{Diag}(\alpha_t) \, S_{t-1} + (1-\alpha_t) v_t^\top$ \\
\midrule
DeltaNet \citep{schlag2021linear} & diag. + low-rank & \ding{51} & $S_t = (I - \beta_t k_t k_t^\top) S_{t-1} + \beta_t k_t v_t^\top$ \\
RWKV-7 \citep{peng2025rwkv7} & diag.\ + low-rank & \ding{51} & $S_t = (\text{Diag}(\alpha_t) - (\vec{\beta_t} \odot\hat{\kappa}) \hat{\kappa}^\top) S_{t-1} + k_t v_t^\top$ \\
GDN \citep{yang2024gated} & diag. + low-rank & \ding{51} & $S_t = (I - \beta_t k_t k_t^\top) \alpha_t S_{t-1} + \beta_t k_t v_t^\top$ \\
KDA \citep{team2025kimi} & diag.\ + low-rank & \ding{51} & $S_t = (I - \beta_t k_t k_t^\top) \text{Diag}(\alpha_t) \, S_{t-1} + \beta_t k_t v_t^\top$ \\
\midrule
\textbf{FG$^2$-GDN} & diag.\ + low-rank & \ding{51} & $S_t = (I - \hat{k}_t \hat{k}_t^\top) \text{Diag}(\alpha_t) \, S_{t-1} + \hat{k}_t \hat{v}_t^\top$ \\
\textbf{FG$^2$-GDN+} & diag.\ + low-rank & \ding{51} & $S_t = (I - \hat{k}_t \hat{k}_t^\top) \text{Diag}(\alpha_t) \, S_{t-1} + \hat{k}_t \tilde{v}_t^\top$ \\
\bottomrule
\end{tabular}%
}
\caption{Unified view of linear recurrent models through state update rules. We classify transition matrices $A_t$ by structure (scalar, diagonal, or low-rank) and whether the write term involves a delta-rule correction.}
\label{tab:update-rules}
\end{table*}

Early gated variants employ scalar or diagonal decay: RetNet \citep{sun2023retentive} uses a fixed scalar $\gamma$; Mamba/Mamba-2 \citep{gu2024mamba, dao2024transformers} introduce data-dependent scalar gates; GLA \citep{yang2024gla} and RWKV-6 \citep{peng2023rwkv} extend to channel-wise diagonal gates $\text{Diag}(\alpha_t)$, providing each feature dimension with an independent forgetting rate. HGRN-2 \citep{qin2024hgrn2} further adopts diagonal gating with a coupled write term.
The most recent RWKV-7 \citep{peng2025rwkv7} goes beyond purely diagonal transitions by introducing a data-dependent low-rank correction, yielding an update $S_t = (\text{Diag}(\alpha_t) - (\vec{\beta_t} \odot\hat{\kappa}) \hat{\kappa}^\top) S_{t-1} + k_t v_t^\top$ that shares structural similarity with the delta rule while augmenting the data-dependent diagonal decay with an untied rank-1 correction. This design also admits an online learning interpretation analogous to DeltaNet's.

\paragraph{Delta Rule and Online Learning.}
The delta rule, also known as the Widrow--Hoff learning rule \citep{widrow1988adaptive}, interprets each recurrent update as a single SGD step on the objective $\|S^\top k_t - v_t\|^2$. DeltaNet \citep{schlag2021linear, yang2024parallelizing} introduced this principle to linear attention, achieving superior associative recall through rank-1 corrective updates. Gated DeltaNet (GDN) \citep{yang2024gated} unified the delta rule with a scalar forget gate $\alpha_t$, and Longhorn \citep{liu2025longhorn} formalized the connection to amortized online learning. KDA \citep{team2025kimi} refined the gate from scalar to channel-wise $\text{Diag}(\alpha_t)$, significantly improving long-context modeling. Our FG$^2$-GDN extends this line of work along the orthogonal axis of \emph{writing granularity}, replacing the scalar learning rate $\beta_t$ with a per-coordinate vector $\beta_t \in \mathbb{R}^{d_k}$.

\paragraph{Richer Test-Time Memory Optimizers.}
Titans \citep{behrouz2025titans} updates a deep neural memory with gradient descent and momentum, while Atlas \citep{behrouz2025atlas} optimizes memory over a local context and uses a Muon-style update that approximates second-order information. These methods enrich the memory architecture and internal optimizer. FG$^2$-GDN instead retains a first-order, matrix-valued delta rule and refines its per-coordinate learning rate while preserving KDA's parallelization path; the directions are therefore complementary.

\paragraph{Hybrid Architectures.}
Purely linear models are fundamentally constrained by finite-state capacity, making exact copying and fine-grained long-context retrieval challenging \citep{gu2024mamba}. Hybrid architectures that interleave linear and full attention layers have emerged as a practical compromise. Jamba \citep{lieber2024jamba} combines Mamba with Transformer blocks in a MoE setting; Gated DeltaNet \citep{yang2024gated} explored hybrids with sliding-window and global attention; Kimi Linear \citep{team2025kimi} adopts a 3:1 KDA-to-MLA ratio, demonstrating that such hybrids can \emph{surpass} full-attention baselines in both quality and efficiency at scale. Our work follows this inter-layer hybrid paradigm, using the same 3:1 ratio to ensure fair comparison, and focuses on improving the linear attention component.

\section{Conclusion}

We present FG$^2$-GDN, a symmetric channel-wise delta-rule parameterization that preserves KDA's tied diagonal-plus-rank-1 structure and specialized chunkwise parallelization path. At 1.3B, an isolating ablation shows that channel-wise writing helps independently of channel-wise decay, and checkpoint analysis confirms that the learned rates remain non-degenerate. FG$^2$-GDN+ progressively adds separate key/value control and achieves the strongest aggregate long-context results, with less than 1\% additional parameters and architectural FLOPs.

\section*{Limitations}

\paragraph{Scale and Architecture Coverage.}
Our experiments are conducted at the 340M and 1.3B parameter scales with dense hybrid architectures. Each configuration is pretrained once under a shared controlled setup, so we cannot report across-seed variance or performance after method-specific hyperparameter tuning. Due to computational resource constraints, we have not validated FG$^2$-GDN on larger-scale models (e.g., 7B or above) or Mixture-of-Experts architectures.

\paragraph{Checkpoint Adaptation.}
Our models are pretrained from scratch, and we have not evaluated adaptation from an existing KDA checkpoint. A possible lower-cost route is to initialize the new channel-wise $\beta$ pathway to reproduce the original scalar rate and then fine-tune the expanded model; evaluating this warm-start strategy remains future work.

\paragraph{Inference Challenges.}
While we demonstrate that FG$^2$-GDN preserves favorable training efficiency, deploying linear-attention hybrids in existing inference systems presents engineering challenges that await further investigation. Current serving frameworks are heavily optimized around the KV-cache paradigm of softmax attention, whereas linear attention maintains a fixed-size recurrent state ($d_k \times d_v$ per head) that requires different cache management strategies. Moreover, speculative decoding requires profiling heterogeneous memory-access patterns across layer types and rolling back recurrent state when draft tokens are rejected. Adapting these inference-time optimizations to linear-attention hybrids remains an open problem.

% Bibliography
\bibliography{custom}

\appendix

\section{Experimental Details}
\label{app:exp-details}

All models in our experiments share an identical hybrid architecture that interleaves Multi-head Latent Attention (MLA) \citep{deepseekai2024deepseekv2strongeconomicalefficient} layers with linear attention layers in a 3:1 ratio (i.e., three linear attention layers followed by one MLA layer). This design follows the inter-layer hybrid paradigm adopted by Kimi Linear \citep{team2025kimi} and ensures that the only variable across compared models is the linear attention mechanism itself.

We experiment at two scales: \textbf{340M} parameters (20 layers, hidden size 1024) and \textbf{1.3B} parameters (24 layers, hidden size 2048). Both MLA and linear attention use 16 heads. The MLA layers adopt low-rank projections for queries and key-value pairs as shown in Table~\ref{tab:setting}, and we disable RoPE in MLA layers so that the linear attention layers can serve as the sole source of positional information through their recurrent dynamics, following the design choice in KDA. The linear attention head dimension is fixed at 128 across both scales.

Both scales are trained with the AdamW optimizer \citep{loshchilov2017decoupled} using a cosine annealing schedule with linear warmup. The peak learning rate is $3 \times 10^{-4}$, decaying to $3 \times 10^{-5}$. The 340M model is trained on 15B tokens with 0.5B warmup tokens and a per-step batch size of 0.5M tokens; the 1.3B model is trained on 100B tokens with 1B warmup tokens and 2M tokens per step. All training uses a sequence length of 8192 and BF16 mixed precision.

\begin{table}[t]
\centering
\resizebox{\columnwidth}{!}{%
\begin{tabular}{cc|cc}
\toprule
\multicolumn{2}{c|}{\textbf{Model Architecture}} & \multicolumn{2}{c}{\textbf{Training Hyperparameters}}           \\ \midrule
\textbf{Parameter}     & \textbf{340M / 1.3B}    & \textbf{Hyperparameter} & \textbf{340M / 1.3B}                  \\ \midrule
Number of layers    & 20 / 24     & Optimizer       & AdamW            \\
Hidden size            & 1024 / 2048             & Learning rate           & $3 \times 10^{-4} \to 3 \times 10^{-5}$ \\
Hybrid mode         & 3:1         & LR Scheduler    & Cosine w/ Warmup \\
MLA / Linear heads  & 16          & Warmup tokens   & 0.5B / 1B        \\
MLA q\_lora\_rank   & 384 / 1536  & Training tokens & 15B / 100B       \\
MLA kv\_lora\_rank  & 128 / 512   & Sequence length & 8192             \\
Linear head dim     & 128         & Tokens per step & 0.5M / 2M        \\
Intermediate size   & 3072 / 6144 & Precision       & BF16             \\
Activation function & SiLU        & MLA RoPE        & Off              \\
\bottomrule
\end{tabular}%
}
\caption{Model architecture and training hyperparameters for 340M and 1.3B scales.}
\label{tab:setting}
\end{table}

\end{document}